# Code Similarity on High Level Programs

M. Mirón-Bernal    H. Coyote-Estrada    J. Figueroa-Nazuno
*Centro de Investigación en Computación. Instituto Politécnico Nacional*
*Unidad Profesional "Adolfo López Mateos" -Zacatenco- México, DF, C.p.07738*
amirona06@sagitario.cic.ipn.mx, {coyote, jfn}@cic.ipn.mx

## Abstract

*This paper presents a new approach for code similarity on High Level programs. Our technique is based on Fast Dynamic Time Warping, that builds a warp path or points relation with local restrictions. The source code is represented into Time Series using the operators inside programming languages that makes possible the comparison. This makes possible subsequence detection that represent similar code instructions. In contrast with other code similarity algorithms, we do not make features extraction. The experiments show that two source codes are similar when their respective Time Series are similar.*

## 1. Introduction

Building large software systems implies a lot of code functions and several hours of debugging. To avoid this, developers and programmers usually use code fragments previously written. Sometimes, this code fragments are adapted and inserted into new software. When this practice is for taking advantage of another programmer without permission, is called plagiarism or reusability.

On educational institutions with programming classes, the *copy-paste-and-adapt-it* practice is very often. Some of these changes are: variables renaming, data types changed, inserting/deleting comments, alter the instruction sequence; last, requires a deeper understanding of the code to be copied; several times, there is no time for doing this because of delivering times.

Even doing these modifications, the adapted code preserves its *essence* (structure); otherwise, the program probably would not accomplish its purpose.

This paper is organized as follows: On Section 2, we described the techniques and the experimental procedure for code similarity. Furthermore, the results of this experiment are shown, the data test includes 210 C# codes from National Polytechnique Institute programming classes. On Section 3, the conclusions are presented.

## 2. Experimental Methodology

In order to obtain the code structure, mentioned above, we process the source code as follows: First, we did comments and code dependences *(headers)* removal; after that, we replace variable names and literals with pre-defined values. Once we did this, we obtained a transformed code into instruction representation level, this means, we identified the instruction type in the code, some examples are: *{assignment, arithmetic, relational, increment/decrement, indexing, function calls} operations*. Each of these categories has different operators from a high-level language programming.

To make this identification possible, we divided the source code into instructions. Each of these instructions has a real-valued $\theta \in \mathbf{R}$. (More details on Section 2.3). By doing this, we obtained a transform that preserves the *essence* (symbolic representation of the original source code into time series) of the instruction. These instructions are ordered as the original code.

Before we go into the algorithms, some definitions are needed.

**Definition 1:** *Sequence*
$$Q = \{ q_1, q_2, q_3 ... q_n \}, \forall q \in R \quad (1)$$

**Definition 2:** *Euclidean Distance.* Given two sequences $Q = \{q_1, ..., q_n\}$ and $C = \{c_1, ..., c_m\}$, with $n=m$, the Euclidean Distance $D_{euc}$ is defined as follows:

$$D_{euc}(Q,C) = \sqrt{\sum_{i=1}^{n}(q_i - c_i)^2} \quad (2)$$

Now we are ready to go into the algorithms.

## 2.1. Fast Dynamic Time Warping

It is an algorithm introduced by Jang[10] in 2000. FDTW minimizes a base distance (5) between two sequences $Q$ and $C$ (1) of size $|Q|$, $|C|$[1] respectively, i.e., $FDTW(Q,C) \leq D_{base}$. Sometimes, FDTW is used to generate a time "alignment" given two sequences, which means, that FDTW builds a point relation called *warp path*, where the *i-th* point of $Q$, is mapped with the *j-th* point of $C$.

The properties of FDTW are presented below:
$FDTW(\langle\rangle, \langle\rangle) = 0$ (3)
$FDTW(Q, \langle\rangle) = FDTW(\langle\rangle,C) = \infty$ (4)
$D_{base}(q_i, c_j) = \sqrt{(q_i - c_j)^2}$ (5)

FDTW distance is obtained using the algorithm:

```
FDTW (Q,C) Algorithm
//Obtain the base distance for the first elements
1.  M[0,0] = D_base (q_0, c_0)
//Compute the distances for the first row
2.  M[0,j] = M[0,j-1] + D_base (q_0, c_j)
//Compute the distances for the first column
3.  M[i,0] = M[i-1,0] + D_base (q_i, c_0)
//Compute the rest of the cells in the matrix
4.  M[i,j] = D_base (q_i, c_j) + min{ M[i-1,j-1], M[i-1,j], M[i,j-1]}
```

By using this algorithm, we obtain:
$FDTW(Q,C) \to \infty$ when Q and C are different
$FDTW(Q,C) = 0$ when Q and C are equal
$FDTW(Q,C) = FDTW(C,Q)$

**2.1.1. Warp path.** The *warp path* (*W*), is a bi-dimensional array, which contains the points relation (mapping) between two sequences $Q$ and $C$. Where each *k-th* element of *W* is defined as:

---
[1] |.| Denotes the number of the elements contained in the sequence.

$W = \{ w_1, w_2, w_k, ..., w_K \}$
$max(|Q|,|C|) \leq K \leq |Q|+|C|-1$ (6)
$w_k = (i,j), 1 \leq i \leq |Q|, 1 \leq j \leq |C|$

The warp path is subjected to the following constraints:
1. **Begin / End:** This constraint requires that the $q_1$ element must be mapped with the $c_1$ element, and the $q_n$ element must be mapped with the $c_m$ element.
2. **Monotonicity:** The elements in the *warp path* must follow the next condition: $w_{k-1} \leq w_k$ within their respective $(i,j)$ index.
3. **Local Restrictions:** The local restrictions are used to reduce the range search of the elements *w* using the neighborhood of a matrix cell.

On Fig. 1, a FDTW example is presented, using two input sequences, and its corresponding *warp path*.

**Figure 1.** FDTW distances matrix. The cells with red border shows the warp path, and the (9,9) element of the matrix give us the FDTW distance equals to 3.

## 2.2. Subsequences Detection

**Definition 3.** *Subsequence.* The contigue elements of a subsequence, are defined as follows:
$sub(Q) = \{ s_1, s_2, s_3 ... s_z \}, \forall s \in Q$ (7)

On Fig. 1, we presented the FDTW similarity between two sequences obtaining $FDTW(Q,C)=3$; the warp path elements are shown with red border. The main diagonal is marked with the symbol $\triangle$; when the warp path elements and the main diagonal elements are the same we obtain $FDTW(Q,C)=0$, i.e., the input sequences are equal.

We base our idea on the following reasoning. The similarity degree between the subsequences is given by the relationship between the warp path and the main diagonal. Those elements which are included in the warp path and are parallel to the main diagonal, will keep a similarity degree; the closer to the main diagonal the bigger would be their similarity.

Our technique has been tested with Time Series, [3] obtaining the expected results, similar subsequence detection using an automatic no supervised algorithm and make no features extraction.

## 2.3. Source Code Transform

Now we explain the representation transform that we applied to the source codes in order to obtain its sequence representation (1).

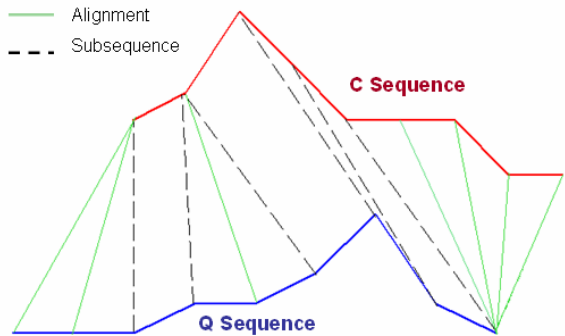

**Figure 2.** Temporal alignment between two sequences Q,C. subsequence identification.

In order to obtain the source code sequence representation, we used the operator category of a language programming and the reserved words of the same language. On Fig. 3, we shown the operators used in the experiments.

For each operators category $\beta$, we gave it a unique value $\phi \in R$. Each operator $\alpha$ within the category $\beta$ we assigned a unique value $\theta \in R$, closer to $\phi$. The value $\theta$ of the operators of each category $\beta$ has the same distance between the elements of the category $\beta$, the same rule applies to the values $\phi$ between categories. By doing this, we obtained two different representations of one source code: in first level, we reemplaze the operators involved in the instruction by the value $\phi$ of the category which contains the operator; in the second level, we reemplaze the same operators using the $\theta$ values, furthermore, we included the respective values of the reserved words, previously assigned $\sigma \in R$.

| Operador category | Operators |
|---|---|
| Arithmetic | + - * / % |
| Logical | & \| ^ ! ~ && \|\| true false |
| Increment, decrement | ++ -- |
| Shift | << >> |
| Relational | == != < > <= >= |
| Assignment | = += -= *= /= % &= \|= ^= <<= >>= |
| Indexing | [ ] |
| Cast | ( ) |
| Object creation | new |
| Type information | as is sizeof typeof |
| Error control | cheked uncheked |

**Figure 3.** Operators category used in the experiments, this operators belong to C# language programming.

Having in mind the previous rules, the intersection of the values $\phi \cap \theta \cap \sigma = \{\}$ [2]. This avoids ambiguity between the operators category, the operators, and the reserved words.

Using the first level representation (operators category), we will obtain the essential structure of instructions. Using a second level representation (operators and reserved words) we obtain the essential structure with more details.

We present some results of this experiment.

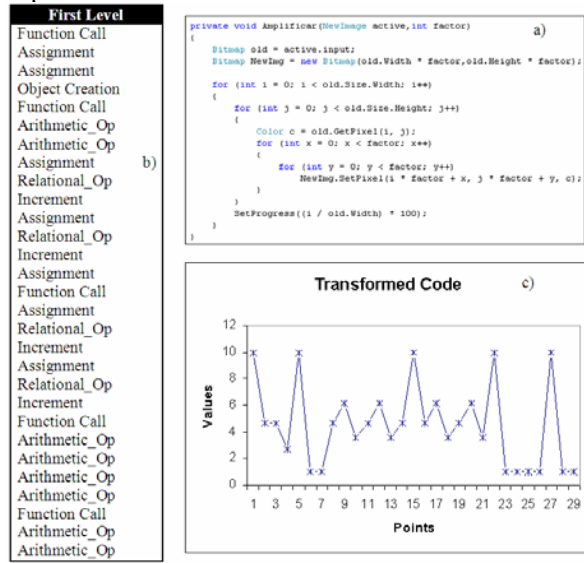

**Figure 4.** Input source code (a), first level representation (b), sequence obtained(c).

---

[2] {.} Denotes the empty set.

## 2.4. Results

The data set contains C# source codes from programming classes of the National Polytechnique Institute. These codes were modified by : reemplazing variable names, data types, alter the instruction sequence order, for mention some of them. By making these systematic modifications we obtained a reference data set, which are similar to each source code from the original data set. The input source codes to our method are free of syntax errors. On figure 5 and 6, we show some of the experiments using a first level representation (operators category).

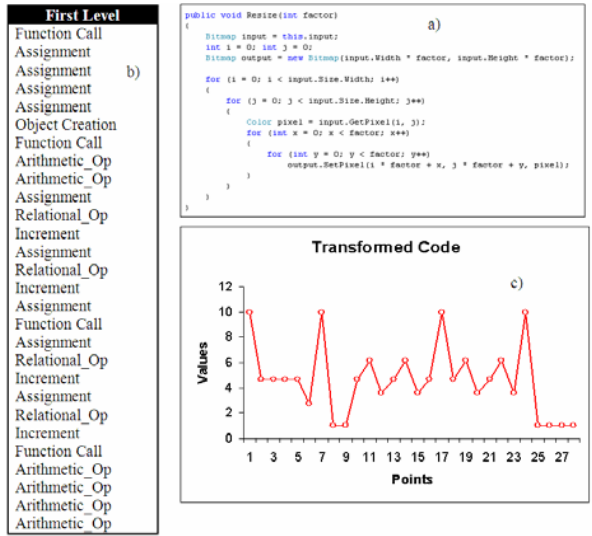

**Figure 5.** Input source code similar to the code shown on fig.4 (a), first level representation (b), sequence obtained(c).

On Figure 7, we show the same source codes but this time, we used the second level representation (operators and reserved words), in this example *FDTW(Q,C)=43.95*.

On Figure 9, we show two source codes with different purpose using first level representation. In this example, there is an evident difference supported by the technique shown above. On Figure 9, *FDTW(Q,C) = 72.50*.

The results suggest that first representation level is used for obtain the essential structure of the code. If the FDTW distance is close to zero, we can transform the source code using the second level representation in order to obtain a sequence with more details. This, helps to make a more detail comparison.

By applying the method presented above, we obtain a similarity value between two source codes. When we use a data set in order to compute similarity among source codes, using our method, we obtain a similarity percentage value, taking in count the maximum and minimum FDTW distance given a source code.

On Figure 10, we show part of the results obtained from the test data set. The query (input source code) is show as column, obtaining the source codes within data set with the minimum FDTW distance, these would be the most similar source codes. For example, if the query is the source code *AMP04* (Fig.10) , the three similar codes with the minimum FDTW distance are: *AMP09, AMP03, AMP05*, with distances 12.80, 15.65 , 15.80 respectively. When we take the maximum distance of the source code data given a source code input, we obtain a similarity percentage. This means, the code *AMP04,* has a similarity degree of 97.43 % with code *AMP09*. The similarity degree between *ED02* and *ED03* (Fig. 10) is 97.14 %. Unlike *ED02* and *AMP06* their similarity degree is 49.19% .

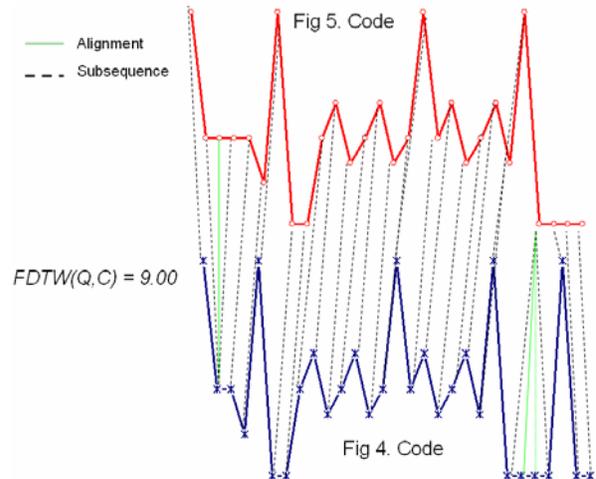

**Figure 6.** Similar subsequences detection, using a first level representation, between two source codes.

## 3. Conclusions

We present a new approach to similarity problem among source codes, without make features extraction.

The subsequence detection method can be applied to any phenomenon that can be represented into Time Series.

The search of new and different representation forms of the entities for similarity comparison, allows the use of several methods and techniques from different specialty areas for computing similarity.

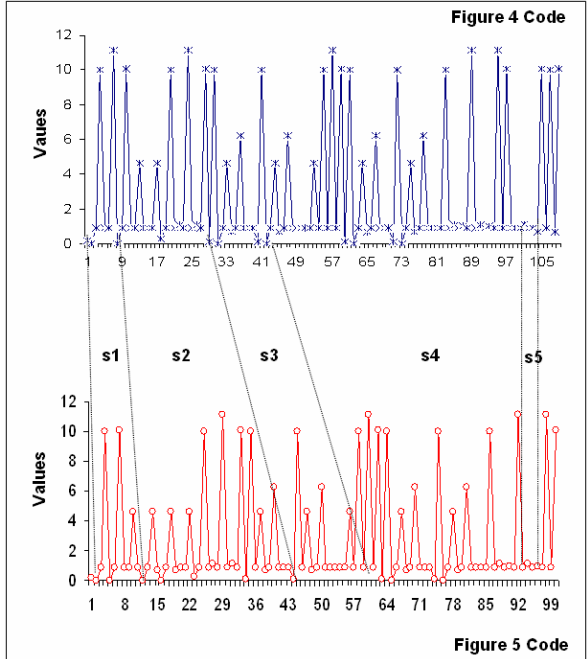

**Figure 7.** Similar subsequence detection, using a second level representation (operators and reserved words) FDTW(Q,C)= 43.95.

```
private void OpenRawImage()
{
    FileStream fs = new FileStream(file, FileMode.Open, FileAccess.Read);
    BinaryReader r = new BinaryReader(fs);
    byte[] temp = new byte[fs.Length];
    for (int i = 0; i < fs.Length; i++)
        temp[i] = r.ReadByte();

    r.Close();
    fs.Close();

    int RawWidth, RawHeight;
    RawWidth = Convert.ToInt32(txtWidth.Text);
    RawHeight = Convert.ToInt32(txtHeigth.Text);
    Bitmap input = new Bitmap(RawWidth, RawHeight, PixelFormat.Format8bppIndexed);
    unsafe
    {
        BitmapData bmData = input.LockBits(
                new Rectangle(0, 0, input.Width, input.Height),
                ImageLockMode.WriteOnly,
                PixelFormat.Format8bppIndexed);
        byte* p = (byte*)bmData.Scan0;
        int diff = bmData.Stride - input.Width;

        for (int i = 0; i < temp.Length; i++)
        {
            *p++ = temp[i];
            if (i % input.Width == 0 && i > 0)
                p += diff;
        }
        input.UnlockBits(bmData);
    }
}
```

**Figure 8.** Source code *RAW01*. The purpose of this code is different from the codes presented above on figures 4 and 5.

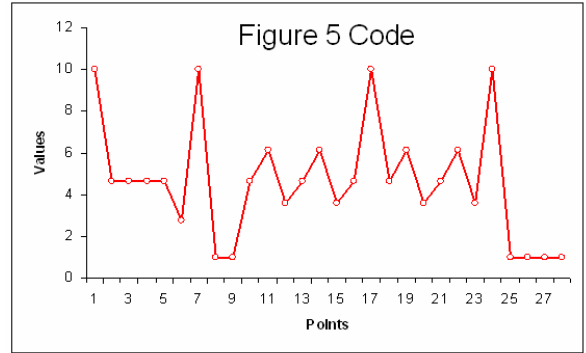

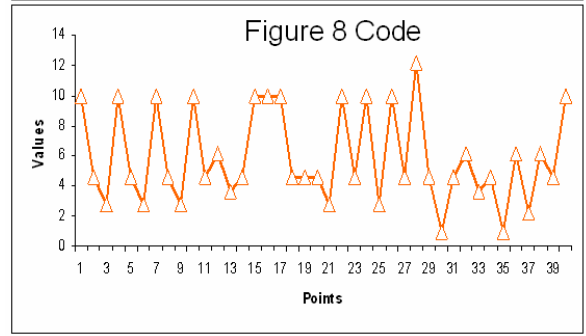

**Figure 9.** Source code similarity comparison using a first level representation. FDTW(Q,C) = 72.50

| | AMP04 | | ED02 | | DMA01 |
|---|---|---|---|---|---|
| AMP04 | 0.00 | ED02 | 0.00 | DMA01 | 0.00 |
| AMP09 | 12.80 | ED03 | 10.50 | DMA02 | 11.30 |
| AMP03 | 15.65 | ED01 | 12.15 | DMA03 | 47.00 |
| AMP05 | 15.80 | ED04 | 22.65 | DMA04 | 66.95 |
| AMP06 | 15.80 | RSC01 | 66.70 | ED01 | 343.90 |
| AMP08 | 28.60 | RAW01 | 95.00 | ED04 | 355.45 |
| AMP11 | 34.05 | RAW02 | 95.00 | ED02 | 355.55 |
| AMP02 | 39.40 | RD01 | 107.35 | ED03 | 367.10 |
| AMP07 | 39.40 | GS01 | 107.40 | RSC01 | 387.40 |
| AMP10 | 39.40 | BW01 | 108.60 | RAW01 | 406.20 |
| HG01 | 46.25 | AMP01 | 114.00 | RAW02 | 406.20 |
| AMP01 | 53.25 | GS02 | 117.45 | RD01 | 411.10 |
| BW01 | 58.60 | HG01 | 123.50 | BW01 | 448.20 |
| GS02 | 76.80 | AMP09 | 137.60 | AMP01 | 450.55 |
| GS01 | 90.45 | AMP08 | 155.95 | GS01 | 459.10 |
| RAW01 | 92.90 | AMP02 | 162.75 | GS02 | 463.70 |
| RAW02 | 118.50 | AMP07 | 162.75 | AMP02 | 469.80 |
| RSC01 | 121.95 | AMP10 | 162.75 | AMP07 | 469.80 |
| RD01 | 135.75 | AMP04 | 166.80 | AMP10 | 469.80 |
| ED03 | 157.35 | AMP11 | 167.05 | HG01 | 469.95 |
| ED04 | 158.85 | AMP03 | 174.90 | AMP11 | 474.05 |
| ED02 | 166.80 | AMP05 | 186.65 | AMP09 | 479.30 |
| ED01 | 169.35 | AMP06 | 186.65 | AMP03 | 483.80 |
| DMA04 | 436.15 | DMA04 | 327.80 | AMP04 | 486.50 |
| DMA01 | 486.50 | DMA01 | 355.55 | AMP08 | 507.25 |
| DMA03 | 492.20 | DMA03 | 364.80 | AMP05 | 511.35 |
| DMA02 | 499.35 | DMA02 | 367.40 | AMP06 | 511.35 |

**Figure 10.** FDTW distances table from the test data set, using a first level representation. The rows within a red border, are the three source codes more similar to the input source code.